%% file: ms.tex
\documentclass[11pt,a4paper]{article}
\usepackage{authblk}
\usepackage[hyperref]{acl2017}
\usepackage{times}
\usepackage{latexsym}
\usepackage{bm}
\usepackage{url}
\usepackage{booktabs}
\usepackage{multirow}
\usepackage{array}
\usepackage{graphicx}
\usepackage{caption}
\usepackage{subcaption}

\newcolumntype{C}[1]{>{\centering\let\newline\\\arraybackslash\hspace{0pt}}m{#1}}

\usepackage[usenames,dvipsnames]{pstricks}
\usepackage{epsfig}
\usepackage{pst-grad} 
\usepackage{pst-plot} 
\usepackage[space]{grffile} 
\usepackage{etoolbox} 
\makeatletter 
\patchcmd\Gread@eps{\@inputcheck#1 }{\@inputcheck"#1"\relax}{}{}
\makeatother

\aclfinalcopy 

\title{Improving Language Modeling using\\Densely Connected Recurrent Neural Networks}

\author{\textbf{Fr\'{e}deric Godin, Joni Dambre and Wesley De Neve }}
\affil{IDLab - ELIS, Ghent University - imec, Ghent, Belgium\\
\tt{firstname.lastname@ugent.be}}

\date{}

\begin{document}
\maketitle
\begin{abstract}
In this paper, we introduce the novel concept of densely connected layers into recurrent neural networks. We evaluate our proposed architecture on the Penn Treebank language modeling task. We show that we can obtain similar perplexity scores with six times fewer parameters compared to a standard stacked 2-layer LSTM model trained with dropout \cite{Zaremba}. In contrast with the current usage of skip connections, we show that densely connecting only a few stacked layers with skip connections already yields significant perplexity reductions.


\end{abstract}

\section{Introduction}
Language modeling is a key task in Natural Language Processing (NLP), lying at the root of many NLP applications such as syntactic parsing \cite{ling}, machine translation \cite{Sutskever} and speech processing \cite{irie2016lstm}. 


In \citet{MikolovKBCK10}, recurrent neural networks were first introduced for language modeling. Since then, a number of improvements have been proposed. \citet{Zaremba} used a stack of Long Short-Term Memory (LSTM) layers trained with dropout applied on the outputs of every layer, while \citet{Gal2016Theoretically} and \citet{DBLP:journals/corr/InanKS16} further improved the perplexity score using variational dropout. Other improvements are more specific to language modeling, such as adding an extra memory component \cite{pointersentinel} or tying the input and output embeddings \cite{DBLP:journals/corr/InanKS16,DBLP:journals/corr/PressW16}.

To be able to train larger stacks of LSTM layers, typically four layers or more \cite{googletranslation}, skip or residual connections are needed. \citet{googletranslation} used residual connections to train a machine translation model with eight LSTM layers, while \citet{pixelrnn} used both residual and skip connections to train a pixel recurrent neural network with twelve LSTM layers. In both cases, a limited amount of skip/residual connections was introduced to improve gradient flow.

In contrast, \citet{dense} showed that densely connecting more than 50 convolutional layers substantially improves the image classification accuracy over regular convolutional and residual neural networks. More specifically, they introduced skip connections between \emph{every} input and \emph{every} output of \emph{every} layer.

Hence, this motivates us to densely connect all layers within a stacked LSTM model using skip connections between every pair of layers. 

In this paper, we investigate the usage of skip connections when stacking multiple LSTM layers in the context of language modeling. When every input of every layer is connected with every output of every other layer, we get a densely connected recurrent neural network. In contrast with the current usage of skip connections, we demonstrate that skip connections significantly improve performance when stacking only a \emph{few} layers. Moreover, we show that densely connected LSTMs need fewer parameters than stacked LSTMs to achieve similar perplexity scores in language modeling.



\section{Background: Language Modeling}
A language model is a function, or an algorithm for learning such a function, that captures the salient statistical characteristics of sequences of words in a natural language. It typically allows one to make probabilistic predictions of the next word, given preceding words \cite{Bengio:2008}. 
Hence, given a sequence of words $[w_1,...w_T]$, the goal is to estimate the following joint probability:
\begin{equation}
Pr(w_1,...,w_T) = \displaystyle\prod_{t=1}^{T}{Pr(w_t|w_{t-1},..w_1)}
\end{equation}
In practice, we try to minimize the negative log-likelihood of a sequence of words:
\begin{equation}
L_{neg}(\theta)= -\displaystyle\sum_{t=1}^{T}{log(Pr(w_t|w_{t-1},..w_1))}.
\end{equation}
Finally, perplexity is used to evaluate the performance of the model:
\begin{equation}
Perplexity = \exp\left(\frac{L_{neg}(\theta)}{T}\right)
\label{eq:perplexity}
\end{equation}

\section{Methodology}
Language Models (LM) in which a Recurrent Neural Network (RNN) is used are called Recurrent Neural Network Language Models (RNNLMs) \cite{MikolovKBCK10}. Although there are many types of RNNs, the recurrent step can formally be written as:
\begin{equation}
\bm{h_t} = f_{\theta}(\bm{x_t},\bm{h_{t-1}})
\end{equation}
in which $\bm{x_t}$ and $\bm{h_t}$ are the input and the hidden state at time step $t$, respectively. The function $f_{\theta}$ can be a basic recurrent cell, a Gated Recurrent Unit (GRU), a Long Short Term Memory (LSTM) cell, or a variant thereof. 

The final prediction $Pr(w_t|w_{t-1},..w_1)$ is done using a simple fully connected layer with a softmax activation function:
\begin{equation}
\bm{y_t} = softmax(\bm{W h_t}+\bm{b}).
\end{equation}

\begin{figure*}[htb]
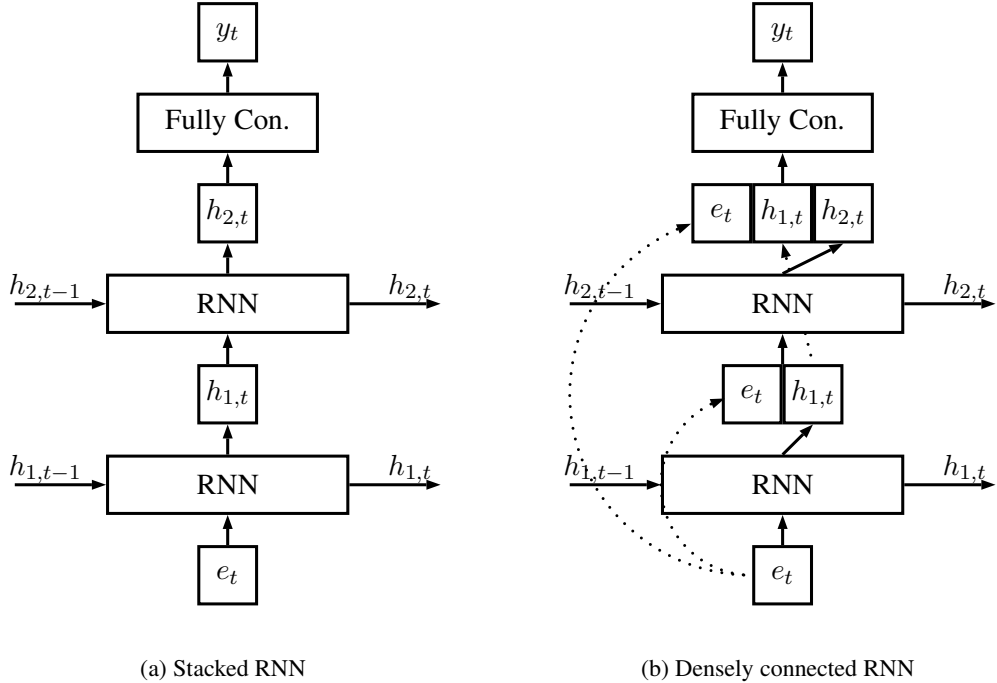

    \centering
    \begin{subfigure}[b]{0.45\textwidth}
    	\centering
        \include{figure1}
        \caption{Stacked RNN}
        \label{fig:stackedrnn}
    \end{subfigure}
    \begin{subfigure}[b]{0.45\textwidth}
    	\centering
        \include{figure2}
        \caption{Densely connected RNN}
        \label{fig:densernn}
    \end{subfigure}
    \caption{Example architecture for a model with two recurrent layers.}\label{fig:architecture}
\end{figure*}

\paragraph{Stacking multiple RNN layers}
To improve performance, it is common to stack multiple recurrent layers. To that end, the hidden state of a a layer $l$ is used as an input for the next layer $l+1$. Hence, the hidden state $\bm{h_{l,t}}$ at time step $t$ of layer $l$ is calculated as:
\begin{equation}
\bm{x_{l,t}} = \bm{h_{l-1,t}},
\end{equation}
\begin{equation}\label{eq:recurrentstep}
\bm{h_{l,t}} = f_{\theta_l}(\bm{x_{l,t}},\bm{h_{l,t-1}}).
\end{equation}
An example of a two-layer stacked recurrent neural network is illustrated in Figure~\ref{fig:stackedrnn}.


However, stacking too many layers obstructs fluent backpropagation. Therefore, skip connections or residual connections are often added. The latter is in most cases a way to avoid increasing the size of the input of a certain layer (i.e., the inputs are summed instead of concatenated).

A skip connection can be defined as:
\begin{equation}
\bm{x_{l,t}} = [\bm{h_{l-1,t}};\bm{h_{l-2,t}}]
\end{equation}
while a residual connection is defined as:
\begin{equation}
\bm{x_{l,t}} = \bm{h_{l-1,t}} + \bm{h_{l-2,t}}.
\end{equation}
Here, $\bm{x_{l,t}}$ is the input to the current layer as defined in Equation \ref{eq:recurrentstep}.

\paragraph{Densely connecting multiple RNN layers}
In analogy with DenseNets \cite{dense}, a densely connected set of layers has skip connections from every layer to every other layer. Hence, the input to RNN layer $l$ contains the hidden states of all lower layers at the same time step $t$, including the output of the embedding layer $e_{t}$:
\begin{equation}
\bm{x_{l,t}} = [\bm{h_{l-1,t}};...;\bm{h_{1,t}};\bm{e_{t}}].
\end{equation}

Due to the limited number of RNN layers, there is no need for compression layers, as introduced for convolutional neural networks \cite{dense}. Moreover, allowing the final classification layer to have direct access to the embedding layer showed to be an important advantage. Hence, the final classification layer is defined as:
\begin{equation}
\bm{y_t} = softmax(\bm{W}[\bm{h_{L,t}};...;\bm{h_{1,t}};\bm{e_{t}}]+\bm{b}).
\end{equation}
An example of a two-layer densely connected recurrent neural network is illustrated in Figure~\ref{fig:densernn}.

\section{Experimental Setup}
We evaluate our proposed architecture on the Penn Treebank (PTB) corpus. We adopt the standard train/validation/test splits \cite{mikolovzweig}, containing 930k training, 74k validation, and 82k test words. The vocabulary is limited to 10,000 words. Out-of-vocabulary words are replaced with an UNK token. 

Our baseline is a stacked Long Short-Term Memory (LSTM) network, trained with regular dropout as introduced by \citet{Zaremba}. 
Both the stacked and densely connected LSTM models consist of an embedding layer followed by a variable number of LSTM layers and a single fully connected output layer. While \citet{Zaremba} only report results for two stacked LSTM layers, we also evaluate a model with three stacked LSTM layers, and experiment with two, three, four and five densely connected LSTM layers. 
The hidden state size of the densely connected LSTM layers is either 200 or 650. The size of the embedding layer is always 200. 

We applied standard dropout on the output of every layer. We used a dropout probability of 0.6 for models with size 200 and 0.75 for models with hidden state size 650 to avoid overfitting. Additionally, we also experimented with Variational Dropout (VD) as implemented in \citet{DBLP:journals/corr/InanKS16}. We initialized all our weights uniformly in the interval [-0.05;0.05]. In addition, we used a batch size of 20 and a sequence length of 35 during training. We trained the weights using standard Stochastic Gradient Descent (SGD) with the following learning rate scheme: training for six epochs with a learning rate of one and then applying a decay factor of 0.95 every epoch. We constrained the norm of the gradient to three. We trained for 100 epochs and used early stopping.
The evaluation metric reported is perplexity as defined in Equation~\ref{eq:perplexity}. The number of parameters reported is calculated as the sum of the total amount of weights that reside in every layer.

Note that apart from the exact values of some hyperparameters, the experimental setup is identical to \citet{Zaremba}.

\section{Discussion}
The results of our experiments are depicted in Table~\ref{tab:tab_results}. The first three results, marked with \emph{stacked LSTM \cite{Zaremba}}, follow the setup of \citet{Zaremba} while the other results are obtained following the setup described in the previous section.
\begin{table*}[htbp]
\centering
\caption{Evaluation of densely connected recurrent neural networks for the PTB language modeling task.}
\include{table_results}
\label{tab:tab_results}
\end{table*}

The smallest densely connected model which only uses two LSTM layers and a hidden state size of 200, already reduces the perplexity with 20\% compared to a two-layer stacked LSTM model with a hidden state size of 200. Moreover, increasing the hidden state size to 350 to match the amount of parameters the two-layer densely connected LSTM model contains, does not result in a similar perplexity. The small densely connected model still realizes a 9\% perplexity reduction with an equal amount of parameters.  

When comparing with \citet{Zaremba}, the smallest densely connected model outperforms the stacked LSTM model with a hidden state size of 650. Moreover, adding one additional layer is enough to obtain the same perplexity as the best model used in \citet{Zaremba} with a hidden state size of 1500. However, our densely connected LSTM model only uses 11M parameters while the stacked LSTM model needs six times more parameters, namely 66M. Adding a fourth layer further reduces the perplexity to 76.8. 

Increasing the hidden state size is less beneficial compared to adding an additional layer, in terms of parameters used. Moreover, a dropout probability of 0.75 was needed to reach similar perplexity scores. Using variational dropout with a probability of 0.5 allowed us to slightly improve the perplexity score, but did not yield significantly better perplexity scores, as it does in the case of stacked LSTMs \cite{DBLP:journals/corr/InanKS16}. 

In general, adding more parameters by increasing the hidden state and performing subsequent regularization, did not improve the perplexity score. While regularization techniques such as variational dropout help improving the information flow through the layers, densely connected models solve this by adding skip connections. Indeed, the higher LSTM layers and the final classification layer all have direct access to the current input word and corresponding embedding. When simply stacking layers, this embedding information needs to flow through all stacked layers. 
This poses the risk that embedding information will get lost. Increasing the hidden state size of every layer improves information flow. By densely connecting all layers, this issue is mitigated. Outputs of lower layers are directly connected with higher layers, effectuating efficient information flow. 

\footnotetext{There are no results reported in \citet{Zaremba} for a small network with dropout. These are our own results, following the exact same setup as for the medium-sized architecture. }

\paragraph{Comparison to other models}
In Table~\ref{tab:tab_comparison}, we list a number of closely related models. A densely connected LSTM model with an equal number of parameters outperforms a combination of RNN, LDA and Kneser Ney \cite{mikolovzweig}.  Applying Variational Dropout (VD) \cite{DBLP:journals/corr/InanKS16} instead of regular dropout \cite{Zaremba} can further reduce the perplexity score of stacked LSTMs, but does not yield satisfactory results for our densely connected LSTMs. However, a densely connected LSTM with four layers still outperforms a medium-sized VD-LSTM while using fewer parameters. \citet{DBLP:journals/corr/InanKS16} also tie the input and output embedding together (cf. model VD-LSTM+REAL). This is, however, not possible in densely connected recurrent neural networks, given that the input and output embedding layer have different sizes.
\begin{table*}[t]
\centering
\caption{Comparison to other language models evaluated on the PTB corpus.}
\include{table_comparison}
\label{tab:tab_comparison}
\end{table*}

\section{Conclusions}
In this paper, we demonstrated that, by simply adding skip connections between \emph{all} layer pairs of a neural network, we are able to achieve similar perplexity scores as a large stacked LSTM model \cite{Zaremba}, with six times fewer parameters for the task of language modeling. The simplicity of the skip connections allows them to act as an easy add-on for many stacked recurrent neural network architectures, significantly reducing the number of parameters. Increasing the size of the hidden states and variational dropout did not yield better results over small hidden states and regular dropout. In future research, we would like to investigate how to properly regularize larger models to achieve similar perplexity reductions.

\section*{Acknowledgments}


The research activities as described in this paper were funded by Ghent University, imec, Flanders Innovation \& Entrepreneurship (VLAIO), the Fund for Scientific Research-Flanders (FWO-Flanders), and the EU.

\bibliographystyle{acl_natbib}

\input{ms.bbl}

\end{document}

%% file: figure1.tex
\psscalebox{1.0 1.0} 
{
\begin{pspicture}(0,-4.0)(5.5369587,4.0)
\psframe[linecolor=black, linewidth=0.04, dimen=outer](3.23,-0.8)(2.43,-1.6)
\psframe[linecolor=black, linewidth=0.04, dimen=outer](4.43,0.4)(1.23,-0.4)
\rput(2.83,0.0){RNN}
\rput(2.83,-1.2){$h_{1,t}$}
\psline[linecolor=black, linewidth=0.04, arrowsize=0.05291667cm 2.0,arrowlength=1.4,arrowinset=0.0]{->}(2.83,-0.8)(2.83,-0.4)
\psframe[linecolor=black, linewidth=0.04, dimen=outer](4.43,-2.0)(1.23,-2.8)
\psframe[linecolor=black, linewidth=0.04, dimen=outer](3.23,-3.2)(2.43,-4.0)
\psline[linecolor=black, linewidth=0.04, arrowsize=0.05291667cm 2.0,arrowlength=1.4,arrowinset=0.0]{->}(2.83,-3.2)(2.83,-2.8)
\psline[linecolor=black, linewidth=0.04, arrowsize=0.05291667cm 2.0,arrowlength=1.4,arrowinset=0.0]{->}(2.83,-2.0)(2.83,-1.6)
\psline[linecolor=black, linewidth=0.04, arrowsize=0.05291667cm 2.0,arrowlength=1.4,arrowinset=0.0]{->}(2.83,0.4)(2.83,0.8)
\rput(2.83,-2.4){RNN}
\rput(2.83,-3.6){$e_t$}
\psline[linecolor=black, linewidth=0.04, arrowsize=0.05291667cm 2.0,arrowlength=1.4,arrowinset=0.0]{->}(0.03,-2.4)(1.23,-2.4)
\psline[linecolor=black, linewidth=0.04, arrowsize=0.05291667cm 2.0,arrowlength=1.4,arrowinset=0.0]{->}(4.43,-2.4)(5.63,-2.4)
\psline[linecolor=black, linewidth=0.04, arrowsize=0.05291667cm 2.0,arrowlength=1.4,arrowinset=0.0]{->}(4.43,0.0)(5.63,0.0)
\psline[linecolor=black, linewidth=0.04, arrowsize=0.05291667cm 2.0,arrowlength=1.4,arrowinset=0.0]{->}(0.03,0.0)(1.23,0.0)
\psframe[linecolor=black, linewidth=0.04, dimen=outer](3.23,1.6)(2.43,0.8)
\rput(2.83,1.2){$h_{2,t}$}
\rput[t](5.23,0.4){$h_{2,t}$}
\rput[t](0.43,0.4){$h_{2,t-1}$}
\rput[t](0.43,-2.0){$h_{1,t-1}$}
\rput[t](5.23,-2.0){$h_{1,t}$}
\psframe[linecolor=black, linewidth=0.04, dimen=outer](4.03,2.8)(1.63,2.0)
\psframe[linecolor=black, linewidth=0.04, dimen=outer](3.23,4.0)(2.43,3.2)
\psline[linecolor=black, linewidth=0.04, arrowsize=0.05291667cm 2.0,arrowlength=1.4,arrowinset=0.0]{->}(2.83,2.8)(2.83,3.2)
\psline[linecolor=black, linewidth=0.04, arrowsize=0.05291667cm 2.0,arrowlength=1.4,arrowinset=0.0]{->}(2.83,1.6)(2.83,2.0)
\rput(2.83,3.6){$y_t$}
\rput(2.83,2.4){Fully Con.}
\end{pspicture}
}

%% file: figure2.tex
\psscalebox{1.0 1.0} 
{
\begin{pspicture}(0,-4.0)(5.5369587,4.0)
\psline[linecolor=black, linewidth=0.04, linestyle=dotted, dotsep=0.10583334cm, arrowsize=0.05291667cm 2.0,arrowlength=1.4,arrowinset=0.0]{<-}(2.83,0.8)(3.23,-0.8)(3.23,-0.8)
\psframe[linecolor=black, linewidth=0.04, dimen=outer](3.63,-0.8)(2.83,-1.6)
\rput(2.83,0.0){RNN}
\rput(3.23,-1.2){$h_{1,t}$}
\psline[linecolor=black, linewidth=0.04, arrowsize=0.05291667cm 2.0,arrowlength=1.4,arrowinset=0.0]{->}(2.83,-0.8)(2.83,-0.4)
\psframe[linecolor=black, linewidth=0.04, dimen=outer](3.23,-3.2)(2.43,-4.0)
\psline[linecolor=black, linewidth=0.04, arrowsize=0.05291667cm 2.0,arrowlength=1.4,arrowinset=0.0]{->}(2.83,-3.2)(2.83,-2.8)
\rput(2.83,-2.4){RNN}
\rput(2.83,-3.6){$e_t$}
\psline[linecolor=black, linewidth=0.04, arrowsize=0.05291667cm 2.0,arrowlength=1.4,arrowinset=0.0]{->}(0.03,-2.4)(1.23,-2.4)
\psline[linecolor=black, linewidth=0.04, arrowsize=0.05291667cm 2.0,arrowlength=1.4,arrowinset=0.0]{->}(4.43,-2.4)(5.63,-2.4)
\psline[linecolor=black, linewidth=0.04, arrowsize=0.05291667cm 2.0,arrowlength=1.4,arrowinset=0.0]{->}(4.43,0.0)(5.63,0.0)
\psline[linecolor=black, linewidth=0.04, arrowsize=0.05291667cm 2.0,arrowlength=1.4,arrowinset=0.0]{->}(0.03,0.0)(1.23,0.0)
\rput(3.63,1.2){$h_{2,t}$}
\rput[t](5.23,0.4){$h_{2,t}$}
\rput[t](0.43,0.4){$h_{2,t-1}$}
\rput[t](0.43,-2.0){$h_{1,t-1}$}
\rput[t](5.23,-2.0){$h_{1,t}$}
\psframe[linecolor=black, linewidth=0.04, dimen=outer](4.03,2.8)(1.63,2.0)
\psframe[linecolor=black, linewidth=0.04, dimen=outer](3.23,4.0)(2.43,3.2)
\psline[linecolor=black, linewidth=0.04, arrowsize=0.05291667cm 2.0,arrowlength=1.4,arrowinset=0.0]{->}(2.83,2.8)(2.83,3.2)
\psline[linecolor=black, linewidth=0.04, arrowsize=0.05291667cm 2.0,arrowlength=1.4,arrowinset=0.0]{->}(2.83,1.6)(2.83,2.0)
\rput(2.83,3.6){$y_t$}
\rput(2.83,2.4){Fully Con.}
\psframe[linecolor=black, linewidth=0.04, dimen=outer](2.83,-0.8)(2.03,-1.6)
\rput(2.43,-1.2){$e_t$}
\psframe[linecolor=black, linewidth=0.04, dimen=outer](3.23,1.6)(2.43,0.8)
\rput(2.83,1.2){$h_{1,t}$}
\psframe[linecolor=black, linewidth=0.04, dimen=outer](2.43,1.6)(1.63,0.8)
\rput(2.03,1.2){$e_t$}
\psline[linecolor=black, linewidth=0.04, arrowsize=0.05291667cm 2.0,arrowlength=1.4,arrowinset=0.0]{->}(2.83,-2.0)(3.23,-1.6)
\psframe[linecolor=black, linewidth=0.04, dimen=outer](4.03,1.6)(3.23,0.8)
\psline[linecolor=black, linewidth=0.04, arrowsize=0.05291667cm 2.0,arrowlength=1.4,arrowinset=0.0]{->}(2.83,0.4)(3.63,0.8)
\psframe[linecolor=black, linewidth=0.04, fillstyle=solid, dimen=outer](4.43,-2.0)(1.23,-2.8)
\psframe[linecolor=black, linewidth=0.04, fillstyle=solid, dimen=outer](4.43,0.4)(1.23,-0.4)
\psarc[linecolor=black, linewidth=0.04, linestyle=dotted, dotsep=0.10583334cm, dimen=outer, arrowsize=0.05291667cm 2.0,arrowlength=1.4,arrowinset=0.0]{<-}(2.43,-2.4){1.2}{107.525566}{270.0}
\psarc[linecolor=black, linewidth=0.04, linestyle=dotted, dotsep=0.10583334cm, dimen=outer, arrowsize=0.05291667cm 2.0,arrowlength=1.4,arrowinset=0.0]{<-}(2.43,-1.2){2.4}{109.592285}{270.0}
\rput(2.83,0.0){RNN}
\rput(2.83,-2.4){RNN}
\end{pspicture}
}

%% file: table_results.tex
\begin{tabular}{ cccccc }
\toprule
\textbf{Name} &  \textbf{Hidden state size} &  \textbf{\# Layers} & \textbf{\# Parameters} & \textbf{Valid} & \textbf{Test} \\ \midrule
 \multirow{3}*{\begin{tabular}{c}
 Stacked LSTM \\ \cite{Zaremba}\footnotemark
\end{tabular}}

& 200 &  2 & 5M  & 104.5 & 100.4 \\ 
& 650 &  2 & 20M & 86.2 & 82.7 \\ 
& 1500 &  2 & 66M & 82.2 & 78.4 \\ 
\midrule
 \multirow{3}{*}{Stacked LSTM} 
 & 200 &  2 & 5M  & 105 & 100.9 \\ 
 & 200 &  3 & 5M  & 113 & 108.8 \\ 
 & 350 &  2 & 9M  & 91.5 & 87.9 \\ 
\midrule
 \multirow{4}{*}{Densely Connected LSTM} & 200 &  2 & 9M  & 83.4 & 80.4 \\ 
 & 200 &  3 & 11M  & 81.5 & 78.5 \\ 
 & 200 &  4 & 14M  & 79.2 & 76.8 \\ 
  & 200 &  5 & 17M  & 79.7 & 76.9 \\ 
\midrule
Densely Connected LSTM & 650 &  2 & 23M  & 81.5 & 78.9 \\ 
Dens. Con. LSTM + Var. Dropout & 650 &  2 & 23M  & 81.3 & 78.3 \\
\bottomrule
\end{tabular}

%% file: table_comparison.tex
\begin{tabular}{ cccc }
\toprule
\textbf{Model} &  \textbf{\# Parameters} &  \textbf{Perplexity Test} \\ \midrule
RNN \cite{mikolovzweig}  &    6M &  124.7 \\ 
RNN+LDA+KN-5+Cache \cite{mikolovzweig} & 9M & 92.0 \\
Medium stacked LSTM \cite{Zaremba} &    20M &  82.7 \\ 
\textbf{Densely Connected LSTM (small - 2 layers)} & \textbf{9M}  &  \textbf{80.4} \\ 
Char-CNN \cite{Kim} & 19M & 78.9 \\
\textbf{Densely Connected LSTM (small - 3 layers)} &  \textbf{11M}  &  \textbf{78.5} \\ 
Large stacked LSTM \cite{Zaremba} & 66M &  78.4 \\ 
Medium stacked VD-LSTM  \cite{DBLP:journals/corr/InanKS16} & 20M & 77.7 \\
\textbf{Densely Connected LSTM (small - 4 layers)} & \textbf{14M}  &  \textbf{76.8} \\ 
Large stacked VD-LSTM \cite{DBLP:journals/corr/InanKS16} & 66M & 72.5 \\
Large stacked VD-LSTM + REAL \cite{DBLP:journals/corr/InanKS16} & 51M & 68.5 \\

\bottomrule
\end{tabular}